\documentclass[letterpaper, 10 pt, conference]{ieeeconf}  % Comment this line out
\IEEEoverridecommandlockouts                              % This command is only
\usepackage{dsfont}
\usepackage{mathrsfs}
\usepackage{verbatim}
\usepackage{amsmath}
\usepackage{amsfonts}
\usepackage{bm}
\usepackage{graphicx}
\usepackage{float}

\usepackage{amsthm}
\usepackage{xcolor}
\usepackage[colorinlistoftodos]{todonotes}
\usepackage{mathtools}
\usepackage{cite}
\usepackage{colortbl}
\allowdisplaybreaks
\usepackage{caption}
\usepackage{subcaption}
\usepackage[per-mode=symbol]{siunitx}
\usepackage{tabularx}
\usepackage{booktabs} 
\usepackage{xspace}

\theoremstyle{definition}
\newtheorem{assumption}{Assumption}

\newtheorem{remark}{Remark}

\theoremstyle{plain}

\usepackage{xspace}    % a very handy package
\usepackage[ruled]{algorithm}
\usepackage[noend]{algpseudocode}

\newcommand{\RR}{\mathbb{R}\xspace}  % Real set
\newcommand{\NN}{\mathbb{N}\xspace}  % Natural number set
\newcommand{\III}{\mathcal{I}\xspace}

\newcommand{\CAV}[1]{CAV\textendash \ensuremath{#1}\xspace}
\newcommand{\HDV}[1]{HDV\textendash\ensuremath{#1}\xspace}

\DeclareMathOperator*{\minimize}{minimize}
\newcommand{\eg}{e.g.,\xspace}
\newcommand{\ie}{i.e.,\xspace}

\newcommand{\etal}{\textit{et al}.}
\title{\LARGE \bf Connected and Automated Vehicles in Mixed-Traffic: Learning Human Driver Behavior for Effective On-Ramp Merging}

\author{Nishanth Venkatesh, Viet-Anh Le, Aditya Dave, {\itshape{Student Members, IEEE,}} \\Andreas A. Malikopoulos, {\itshape{Senior Member, IEEE}} 
	\thanks{This research was supported by NSF under Grants CNS-2149520 and CMMI-2219761. }\thanks{
	The authors are with the Department of Mechanical Engineering, University of Delaware, Newark, DE 19716 USA (email: \texttt{\{nish; vietale; adidave; andreas}\}@udel.edu).} }

\begin{document}

\maketitle
\thispagestyle{empty}

\begin{abstract}
Highway merging scenarios featuring mixed traffic conditions pose significant modeling and control challenges for connected and automated vehicles (CAVs) interacting with incoming on-ramp human-driven vehicles (HDVs). In this paper, we present an approach to learn an approximate information state model of \emph{CAV-HDV interactions} for a CAV to maneuver safely during highway merging. In our approach, the CAV learns the behavior of an incoming HDV using approximate information states before generating a control strategy to facilitate merging. First, we validate the efficacy of this framework on real-world data by using it to predict the behavior of an HDV in mixed traffic situations extracted from the Next-Generation Simulation repository. Then, we generate simulation data for HDV-CAV interactions in a highway merging scenario using a standard inverse reinforcement learning approach. Without assuming a prior knowledge of the generating model, we show that our approximate information state model learns to predict the future trajectory of the HDV using only observations. Subsequently, we generate safe control policies for a CAV while merging with HDVs, demonstrating a spectrum of driving behaviors, from aggressive to conservative. We demonstrate the effectiveness of the proposed approach by performing numerical simulations.
\end{abstract}

\section{Introduction}
\label{section:Introduction}
Connected and automated vehicles (CAVs) have the potential to significantly improve  energy efficiency, safety, and comfort \cite{zhao2019enhanced}. Numerous research efforts have focused on how to ensure energy efficiency \cite{Malikopoulos2020}, safety \cite{xiao_decentralized_2021,ChalakiCBF2022}, and traveler comfort \cite{gao_predictive_2019} with 100 \% CAV penetration. We expect a gradual rise in the number of CAVs on the roads, and thus, CAVs must be able to safely maneuver alongside human-driven vehicles (HDVs). The presence of HDVs poses a challenge in the control of CAVs due to the unpredictability of human driving behavior in various situations such as lane-changing, intersections, roundabouts and highway on-ramp merging. This challenge can be addressed by developing approaches to predict human-driving behavior online and dynamically adjusting the trajectory of a CAV. 

%CAVs have to predict possible future behaviors and plan its trajectories accordingly. The way the control actions need to be chosen may require understanding human driver behavior for various traffic scenarios, such as highway on-ramp merging, intersection, roundabouts.

In recent years, significant attention has been directed toward game-theoretic models of interaction and control for vehicles. Liu \etal \hspace{0.5pt} \cite{liu_cooperation-aware_2021} proposed a decision-making algorithm to address pairwise vehicle interaction in merging scenarios based on the leader-follower game. Chandra and Manocha \cite{chandra_gameplan_2022} proposed a robotics-based auction framework for vehicle navigation at un-signalized intersections, roundabouts, and merging scenarios. A game-theoretic model predictive control combined with a moving horizon inverse reinforcement learning and weight adaptation strategies for CAVs interacting with human drivers were proposed in \cite{Le2022CDC,Le2023ACC}. 

Other approaches have focused on predicting human-driving behavior in different situations with deep learning methods that utilize recurrent layers to learn temporal relations in data. % have also gained focus in predicting human driving behavior under various circumstances. 
For example, Altché and Fortelle \cite{altche2017lstm} proposed a neural network architecture based on a long short-term memory (LSTM) to predict the future trajectory of a vehicle moving on a straight road. Park \etal \cite{park2018sequence}  generated the future trajectories of all vehicles surrounding an HDV using an LSTM-based model.
Similarly, Deo and Trivedi used LSTMs for interaction-aware motion prediction of surrounding vehicles on freeways \cite{deo2018multi} and for robustly learning inter-dependencies of vehicle motion in \cite{deo2018convolutional}.
 LSTMs have also been used in \cite{kim2017probabilistic} with an occupancy grid-based framework to consider latent factors in the predicted trajectories, such as road structure, traffic rules, and a driver’s intention.

While the primary goal of most deep learning methods has been to predict the future behavior of HDVs, there is a growing interest in utilizing such predictions to control CAVs during interactions with HDVs. Kherroubi \etal \cite{el2021novel} trained a neural network model to predict the intentions of human drivers to learn a driving strategy using a deep policy gradient approach. Guo \etal \cite{guo2020merging} applied Q-learning to control the change of lanes for CAVs in mixed traffic where the HDVs follow the Gipps' car-following model. Although there has been significant progress in predicting HDV behavior, control approaches that prioritize safety while retaining the ability to generalize across different models of HDV behavior is still an open research question. Establishing a framework that can lead to the design and implementation of learning-enabled \cite{Malikopoulos2022a} CAVs in which safety is ensured with high levels of confidence is timely.
% which merge learning of HDV behaviors with the control of CAVs that prioritizes safety while generalizing across different models of HDV behavior.
%Liu et al. \hspace{0.5pt} in \cite{liu2021efficient} proposed on-ramp merging strategy to coordinate multi-lane traffic using a variation of deep Q-Learning.

In this paper, we focus on this combined problem of learning HDV behavior and utilizing the learned model to generate safe control strategies for a CAV.
To this end, we draw upon ideas in the literature on partially observed reinforcement learning.
Subramanian \etal \cite{subramanian2019approximate, subramanian2022approximate} presented a principled framework to learn state-space representations and compute control strategies for systems with unobserved hidden states.
Such a model, known as an approximate information state model, has been utilized in robotics \cite{yang2022discrete} for system identification followed by control and in medical dead-end identification for reinforcement learning \cite{fatemi2021medical}. Similarly, Dave \etal \cite{Dave2022approx, dave2023approximate} proposed a non-stochastic framework for learning an approximate model and deriving minimax control strategies. The notion of an approximate information state forms the basis of our approach.

%\subsection{contributions}

The main contributions of this paper are as follows. We provide a methodology to learn an on-ramp HDV's behavior in a highway merging scenario and utilize the learned model to generate a prioritized safety strategy for a CAV on the highway. 
To this end, we formulate the CAV's control problem based on a set of desired objectives, including safety and energy efficiency, and implement the following steps: (1) Construct and train a neural network architecture to predict the trajectory of an HDV without prior knowledge of their dynamics. We achieve this by considering the history of observations to be representative of the HDV's driving behavior and compressing the history into an approximate state representation inspired by \cite{subramanian2022approximate}. (2)  Develop an iterative model predictive control (MPC) algorithm which uses the learned approximate information state model to generate a safe and energy-efficient control strategy for the CAV online. (3) Validate the efficacy of our proposed learning framework by successfully predicting vehicle trajectories from the Next Generation Simulation (NGSIM) repository. (4)Validate the efficacy of our combined learning and control approach to generate CAV actions across 5000 simulations on generated data of HDV-CAV interactions, featuring a variety of HDV behaviors from aggressive to conservative.

The remainder of the paper proceeds as follows. In Section \ref{section:Problem Formulation}, we present our problem formulation for the CAV to merge in the presence of an HDV on-ramp. In Section \ref{section:model}, we present preliminaries on approximate information state based model followed by our encoder-decoder model to learn and predict \emph{CAV-HDV interaction} and how to train the neural network architecture. In Section \ref{section:Simulation}, we provide an analysis on how to solve our problem formulation using the learned model and then present numerical simulation results for various merging scenarios. Finally, in Section \ref{section:conclusion}, we draw concluding remarks and discuss ongoing work.
%\subsection{Literature Review}
%\cite{dave2019decentralized}
%\subsection{Contribution of the Paper}

%\textbf{Some relevant papers on mixed-traffic merging:}
%Liu \etal \cite{liu_cooperation-aware_2021} presented a decision making algorithm to address pairwise vehicle interaction in merging scenarios based on the leader-follower game.
%In \cite{chandra_gameplan_2022}, Chandra and Manocha proposed a robotics-based auction framework for vehicle navigation at unsignalized intersections, roundabouts, and merging scenarios.
%Le and Malikopoulos \cite{Le2022CDC,Le2023ACC} combined game-theoretic model predictive control, moving horizon inverse reinforcement learning, and weight adaptation strategies to address the control problem for CAVs while interacting with human drivers.

\begin{comment}
However, there is strong evidence that the full implementation of CAVs will not happen instantly \cite{barnes2017autonomous}. A gradual introduction of CAVs will require a transition from traffic being comprised only of human-driven vehicles (HDVs) to mixed traffic where HDVs and CAVs simultaneously cruising together
\end{comment}

\section{Problem Formulation}\label{section:Problem Formulation}

\begin{figure}[ht]
  \centering
  %\captionsetup{justification=centering}
  \includegraphics[width=\columnwidth]{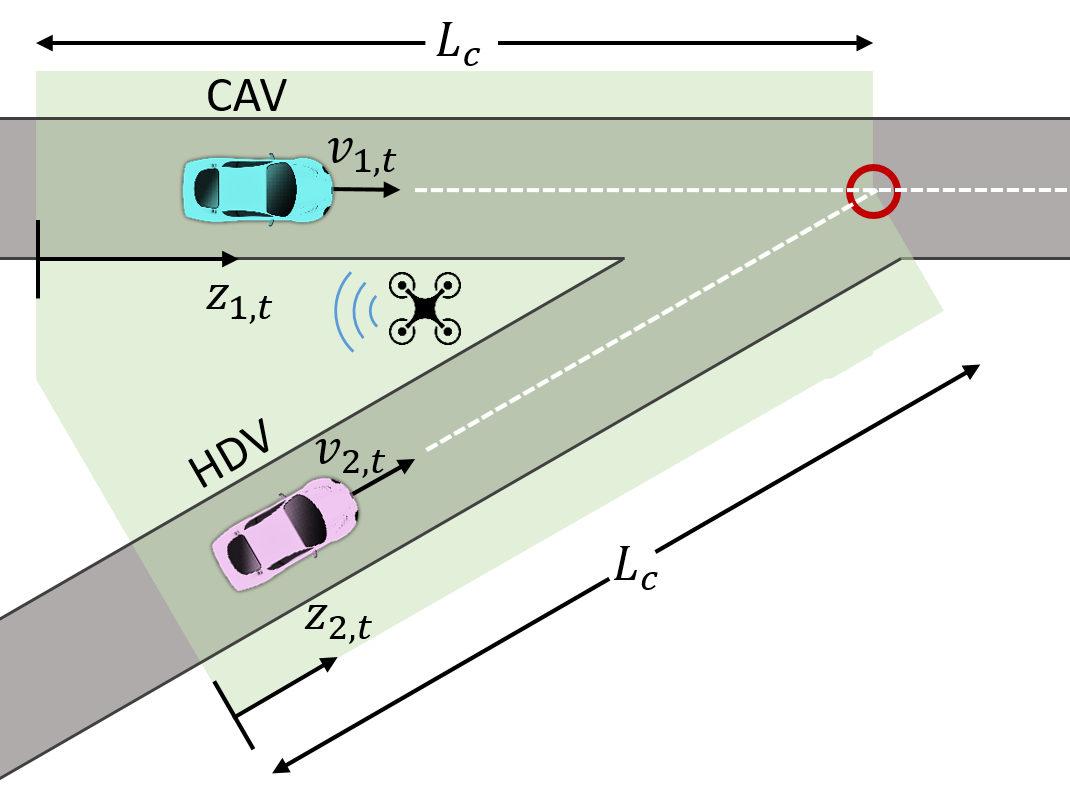}
  \caption{Control zone (green area) of our mixed traffic merging scenario and the conflict point (red circle).}
  \label{fig:simple_mixed_traffic}
\end{figure}

In this section, we formulate an MPC problem to compute the control input for a CAV in a mixed-traffic highway merging scenario.
As shown in Fig. \ref{fig:simple_mixed_traffic}, a CAV indexed by $1$, and an HDV indexed by $2$, travel on a main road and a ramp, respectively. 
We define a \emph{control zone} as highlighted by the green region in Fig. \ref{fig:simple_mixed_traffic}. Within the control zone, the control inputs to \CAV{1} are determined by our proposed method, whereas, outside the control zone, they can be determined by a car-following model, \eg intelligent driver model \cite{treiber2013traffic}.
We consider that the control zone begins at a distance $L_c \in \mathbb{R}_{>0}$ upstream of the \textit{conflict point} on each road, i.e., 
the location where the paths of \CAV{1} and \HDV{2} intersect. The conflict point is the region where a lateral collision may occur, and we mark it by a red circle in Fig. \ref{fig:simple_mixed_traffic}. 
Our goal is to control \CAV{1} to merge safely and effectively, given the presence of \HDV{2} in this merging scenario.

%control_zone

In our formulation, the longitudinal positions of \CAV{1} and \HDV{2} are measured from the beginning of the control zone at any $t\in \mathbb{N}$, we denote them by $z_{1,t} \in \mathbb{R}_{\geq0}$ and $z_{2,t} \in \mathbb{R}_{\geq0}$, respectively. 
Let $v_{1,t} \in \mathbb{R}_{\geq0}$ and $u_{1,t} \in \mathbb{R}$ be the speed and acceleration of \CAV{1}, $v_{2,t} \in \mathbb{R}_{\geq0}$ and $u_{2,t} \in \mathbb{R}$ be the speed and acceleration of \HDV{2} at time $t$. 
We collectively denote the position and speed of each vehicle $i = 1,2$ as its state $x_{i,t} = (z_{i,t} \,, \, v_{i,t})$ at time $t$. 
Then, starting with the initial state $x_{1,0}$ and $x_{2,0}$ at $t=0$, the state of each vehicle evolves according to
\begin{align} \label{dynamics_CAV}
    {x}_{1,t+1}=f_{1,t}({x}_{1,t},{u}_{1,t}), \\
    {x}_{2,t+1}=f_{2,t}({x}_{2,t},{u}_{2,t}).\label{dynamics_HDV}
\end{align}
In this paper, we consider that for all $t \in \mathbb{N}$ the dynamics of \CAV{1} from \eqref{dynamics_CAV} takes the form of the following double-integrator model
\begin{equation}
\label{eq:integrator}
\begin{split}
z_{1,t+1} &= z_{1,t} + \Delta T v_{1,t} + \frac{1}{2} \Delta T^2 u_{1,t} , \\
v_{1,t+1} &= v_{1,t} + \Delta T u_{1,t} , \\
\end{split}
\end{equation}
where $\Delta T \in \mathbb{R}_{>0}$ is the sampling time between two discrete timesteps.
We consider the following state and control constraints for \CAV{1}
\begin{equation}
\label{eq:bound}
0 \le v_{\text{min}} \le v_{1,t} \le v_{\text{max}}, \quad \text{and} \quad u_{\text{min}} \le u_{1,t} \le u_{\text{max}},
\end{equation}
where $u_{\text{min}}$, $u_{\text{max}} \in \RR$ are the minimum deceleration and maximum acceleration, respectively, and $v_{\text{min}}$, $v_{\text{max}} \in \mathbb{R}_{>0}$ are the minimum and maximum speed limits, respectively.
Both \CAV{1} and \HDV{2} are restricted to only move in the forward direction, and hence, their speeds $v_{i,t}$ for $i = 1,2$ are always non-negative. 
However, we do not impose an upper bound on the speed of \HDV{2} that is known a priori because it can be violated by human driving behavior. 
In the next subsection, we formulate the control problem for \CAV{1}.

%%%%%%%%%%%%%%%%%%%%%%%%%%%%%%%%%%%%%%%%%%%%%%
% Viet-Anh is writing this subsection
\subsection{Model Predictive Control Formulation} \label{subsection:mpc_formulation}

In this subsection, we formulate an MPC problem to control \CAV{1} in the merging scenario given the system dynamics \eqref{eq:integrator} and the state and control constraints \eqref{eq:bound}. 
Let $H \in \NN$ be the time length of the control horizon, $t \in \mathbb{N}$ be the current time step, and $\III_t = \{ t, \dots, t+H-1 \}$ be the set of all time steps in the control horizon originating at $t$.
%, $p_{1,k} \in \RR$ is the longitudinal position of the vehicle with respect to the conflict point at time $k$, and  $v_{1,k} \in \RR$ and $a_{1,k} \in \RR$ are the speed and acceleration of the CAV at time $k$, respectively.
%The state and control input of the CAV are defined by $\bb{x}_{1,k} = [p_{1,k}, v_{1,k}]^{\top}$ and $u_{1,k} = a_{1,k}$, respectively.
The objective of the MPC problem can be formed by a linear combination of three features: \textit{(i)} minimizing the control input, \textit{(ii)} minimizing the deviation from the maximum allowed speed to reduce the travel time,
and \textit{(iii)} a logarithmic penalty function corresponding to a collision avoidance constraint \cite{Le2023ACC}.
%Thus, the objective function can be mathematically given by
Thus, the objective function for \CAV{1} at any time $k$ is given by the cost
\begin{equation}
\label{eq:mpc-obj}
\begin{multlined}
l (x_{1,k+1}, u_{1,k})
= \omega_{1} u_{1,k}^2 +
\omega_{2} (v_{1,k+1} - v_{\text{max}})^2 \\
- \omega_{3} \log \big( (z_{1,k+1} - z_c  + \rho v_{1,k+1})^2 \\+ (\hat{z}_{2,k+1} - z_c + \rho \hat{v}_{2,k+1})^2 \big),
\end{multlined}
\end{equation}
where $\omega_{1}$, $\omega_{2}$, and $\omega_{3}$ are positive weights, $z_c$ is the position of the conflict point, and $\rho \in \RR_{>0}$ is a parameter which accounts for the reaction delay of \HDV{2}. An increase in $\rho$ represents an increase in the conservatism of the objective by accounting for \HDV{2} to have greater reaction delays.
In \eqref{eq:mpc-obj}, $\hat{z}_{2,k+1}$ and $\hat{v}_{2,k+1}$ denote the predicted future positions and speeds of \HDV{2} at time step $k+1$. Then, the MPC problem for \CAV{1} is formulated for each $t \in \mathbb{N}$ as
\begin{subequations}
  \label{eq:MPC}
  \begin{align}
    &
    \begin{multlined}
    \underset{ \{u_{1,k}\}_{k \in \III_t} }{\minimize} \; \sum_{k \in \III_t} l ({x}_{1,k+1}, u_{1,k}),
    \end{multlined}
    \label{eq:MPC:obj}\\
    & \text{subject to: } \eqref{eq:integrator}, \eqref{eq:bound}.
    %\nonumber  \\
    %& \quad \text{\eqref{eq:integrator}}, \; \text{\eqref{eq:bound}}.
  \end{align}
\end{subequations}

We seek the control inputs $(u_{1,t}, \dots, u_{1,t+H-1})$ to minimize \eqref{eq:MPC} at each $t \in \mathbb{N}$ without prior knowledge of the dynamics of \HDV{2} given in \eqref{dynamics_HDV}. Thus, \CAV{1} must learn a model that can predict the future trajectory of \HDV{2} required in the objective \eqref{eq:mpc-obj}. To ensure that \CAV{1} has access to the real-time data of \HDV{2}, in our framework, we impose the following assumption.

\begin{assumption} \label{Perfect observation - coordinator}
A coordinator, which we can consider that is a drone shown in Fig. \ref{fig:simple_mixed_traffic}, exists to collect real-time observations on the states and control actions of \HDV{2} and transmit them to \CAV{1} instantaneously. 
The coordinator can achieve this without any noise in either observation or communication.
\end{assumption}

\begin{remark} \label{connecting problem to approach}
The future control actions and the trajectory for \HDV{2} are not pre-determined for the horizon because (i) the dynamics of \HDV{2} are unknown, and (ii) the impact of the state and actions of \CAV{1} on the actions of \HDV{2} is also unknown. Anticipating that \HDV{1} would react in response to the control actions of \CAV{1} \cite{Le2022CDC}, we model the joint dynamics of both \CAV{1} and \HDV{2} as an unknown \emph{partially observable Markov decision process} (POMDP).
Next, we use the framework of approximate information states for POMDPs to learn a model that predicts the trajectory of \HDV{2} coupled with the actions of \CAV{1}. 
\end{remark}

\section{Learning Framework}\label{section:model}

%%%%%%%%%%%%%%%%%%%%%%%%%%%%%%%%%%%%%%%%%%%%%%
% How to cite ours papers in this section
\subsection{ Preliminaries}
In this subsection, we present the mathematical formulation of POMDPs and the framework of approximate information states from \cite{subramanian2019approximate}. This formulation helps us lay the foundation to understand how we analyze our system with incomplete knowledge of underlying dynamics. Subsequently, we will use this framework in our merging scenario to learn a model to predict the future trajectory of \HDV{2}.\\

%%%%%%%% NAme this patch to POMDP
%\begin{remark}\label{POMDP-formal}
\noindent    \textbf{POMDP Fundamentals}:
We define a POMDP as a tuple $( \mathcal{X}, \mathcal{U}, \mathcal{Y}, T, O, c, \gamma)$, where the set of feasible states is $\mathcal{X}$, the set of feasible control actions is $\mathcal{U}$, the set of feasible observations is $\mathcal{Y}$, the function $T: \mathcal{X} \times \mathcal{U} \times \mathcal{X} \to [0,1]$ yields the transition probability $T(x_t,u_{t-1},x_{t+1})= p(x_{t+1}|x_t,u_{t-1})$ for all $x_t, x_{t+1} \in \mathcal{X}$ and $u_t \in \mathcal{U}$, the function $O:\mathcal{X} \times \mathcal{Y} \to [0,1]$ yields the observation probability $O({x}_{t+1},{y}_{t+1})= p({y}_{t+1}|{x}_{t+1})$ for all $x_{t+1} \in \mathcal{X}$ and $y_{t+1} \in \mathcal{Y}$, the function $c: \mathcal{X} \times \mathcal{U} \to \mathbb{R}$ yields the cost $c(x_t,u_{t})$ for all $x_t \in \mathcal{X}$ and $u_t \in \mathcal{U}$, and $\gamma \in [0,1)$ is the discount factor. The history of observations and control actions up to a given time $t$ constitutes the memory defined as ${m}_t=({m}_{t-1},{y}_t,{u}_{t-1}) \in \mathcal{M}_t$.\\

%The memory at time $t$ is a collection of the history of observations and control actions, and is denoted by ${m}_t=({m}_{t-1},{y}_t,{u}_{t-1}) \in \mathcal{M}_t$.\\
%%% Check if my memory update is right 
%\end{remark} 

%\begin{remark}\label{Approximate Information State}
\noindent    \textbf{Approximate Information State}:
% HOw to define the metric space ?
An approximate information state is a compression of the data of our system stored in a memory which can be used to compute an approximately optimal control strategy for a POMDP using a dynamic programming decomposition. In the context of our problem, it has the important property that an approximate information state model can be learned purely from observation data without complete knowledge of underlying dynamics.
Consider a measurable space $(\mathcal{X}, \mathcal{B}(\mathcal{X}))$, where $\mathcal{B}(\mathcal{X})$ is the sigma algebra on $\mathcal{X}$. We denote an integral probability metric between any two probability distributions $\mu, \nu \in \Delta(\mathcal{X})$ by $d(\cdot, \cdot)$. Examples of such metrics are the total variation metric, Wasserstein metric, and maximum mean discrepancy metric.
Then, for a POMDP with a horizon $T \in \mathbb{N}$, a time-invariant approximate information state model is defined by a tuple $(\mathcal{S}, \hat{p}, \hat{c}, \sigma_t: t=0,\dots,T)$, where $\mathcal{S}$ is a Banach space of feasible information states, $\hat{p}: \mathcal{S} \times \mathcal{U} \to \Delta(\mathcal{S})$ is a Markovian probability kernel, $\hat{c}: \mathcal{S} \times \mathcal{U} \to \mathbb{R}$ is an approximate cost function, and $\sigma_t: \mathcal{M}_t \to \mathcal{S}$ is a memory compression function. This constitutes an $(\varepsilon, \delta)$-approximate model of the POMDP if there exist $\varepsilon, \delta \in \mathbb{R}_{>0}$ such that for all realizations of memory ${m}_t \in \mathcal{M}_t$ and realizations of control action ${u}_t \in \mathcal{U}$, the following properties hold:

%at any $t=0,\dots,T$ is a compression of the memory ${s}_t$ which takes values in a space $\mathcal{S}$ and is generated by a compression function $\hat{\sigma}_t: \mathcal{M}_t \to \mathcal{S}$. It is updated in a Markovian fashion such that  $\hat{p}:\mathcal{S} \times \mathcal{U}  \rightarrow \Delta(\mathcal{S})$, and a function to predict the cost $\hat{c}:\mathcal{S} \times \mathcal{U}  \rightarrow \mathbb{R}$. The update equation for the approximate information state is given by ${s}_t=\hat{\sigma}({m}_t)$ and the following properties have to be satisfied for any time $t$, any memory ${m}_t$ and control action ${u}_t$ :\\
% Should I write this in terms of random variables and realizations?  

\noindent    (\textbf{AP1}) Sufficient to predict the cost $c_t=c({x}_t,{u}_t)$ approximately
   \begin{equation}\label{eq:ap_1}
    \left|\mathbb{E}[c_t({x}_t,{u}_t) \, | \, {m}_t,{u}_t]-\hat{c}(\sigma_t({m}_t),{u}_t) \right| \leq \varepsilon.   
   \end{equation}

\noindent    (\textbf{AP2a})  The approximate information state evolves in a deterministic manner, i.e., there exists a measurable update function $\psi: \mathcal{S} \times \mathcal{Y} \times \mathcal{U} \to \mathcal{S}$ such that
   \begin{equation}\label{eq:ap_2a}
    \sigma_{t}({m}_{t})=\psi(\sigma_t({m}_{t-1}),{y}_{t},{u}_{t-1}).
   \end{equation}

\noindent    (\textbf{AP2b})  Sufficient to approximately predict future observations, i.e., there exists a probability kernel $\hat{p}^{\text{y}}: \mathcal{S} \times \mathcal{U} \to \Delta(\mathcal{Y})$ such that for any Borel subset $B$ of $\mathcal{Y}$, the memory-based distribution $\mu_t(B) = p(y_{t+1} \in B|m_t, u_t)$ and the predicted distribution $\phi(B) = \hat{p}^{\text{y}}(y_{t+1} \in B|\sigma_t(m_t), u_t)$ satisfy
   \begin{equation} \label{eq:ap_2b}
    d(\mu_t,\phi) \leq \delta.
   \end{equation}
    
%\end{remark} 

% Is hence the right word after "computationally intensive" ?
An $(\varepsilon, \delta)$-approximate information state model forms a perfectly observed Markov decision process with the state at each $t$ given by $s_t \in \mathcal{S}$. Thus, we can use it to compute an approximately optimal control strategy by formulating a dynamic programming decomposition of the problem. It has been established that such an approximate strategy has a bounded performance loss and that the performance improves linearly with a decrease in $\varepsilon$ and $\delta$ \cite[Theorem 9]{subramanian2019approximate}. This property implies that an approximate information state constitutes a principled compression of the memory, which can be learned by best enforcing $(\textbf{AP1})$ and $(\textbf{AP2b})$ while ensuring that $(\textbf{AP2a})$ is satisfied. 

\begin{figure}[ht!]
  \centering
  %\captionsetup{justification=centering}
  \includegraphics[width=\columnwidth, keepaspectratio]{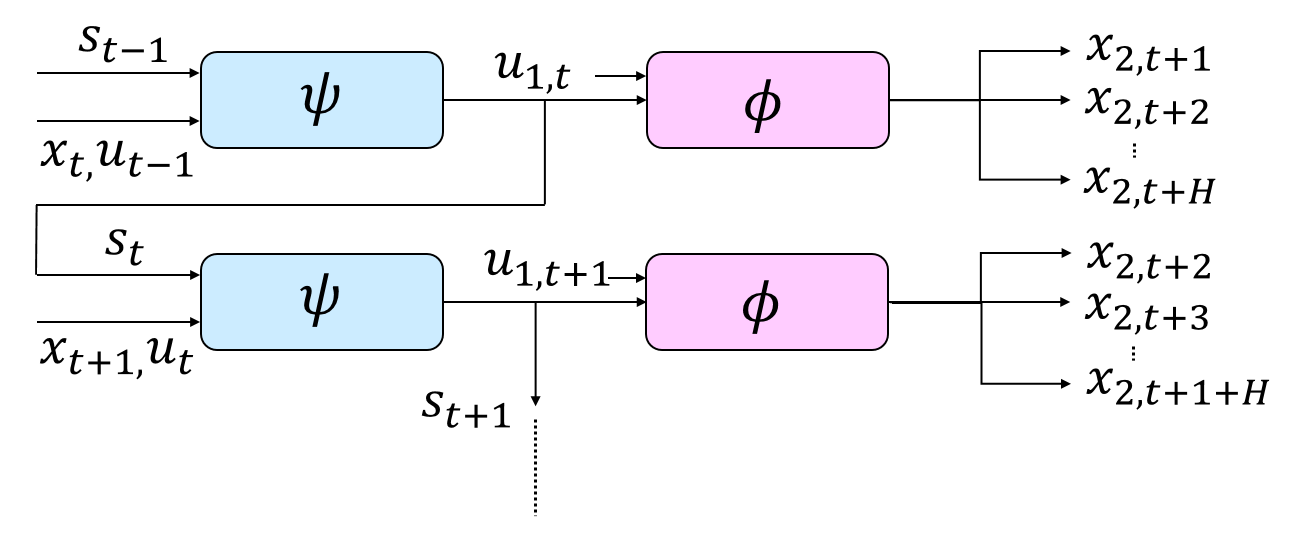}
  \caption{A visualization of the encoder-decoder architecture.}
   \label{fig:encode_decoder_with_action}
   \vspace{-5pt}
\end{figure}

In light of these advantages, we adopt this framework in our CAV-HDV merging scenario to predict the trajectory of \HDV{1} and, subsequently, compute a control strategy for \CAV{1}.
In this framework, at each $t\in \mathbb{N}$, the state of \CAV{1} is $x_{1,t} = (z_{1,t} \,, \, v_{1,t})$, the control action is $u_{1,t}$ and the observation received by \CAV{1} is $y_t = (z_{1,t} \,, \, v_{1,t}\,, \, u_{1, t-1}\,, \,z_{2,t}\,, \,v_{2,t}\,, \,u_{2,t-1})$.
We deploy an encoder-decoder neural network architecture to represent an approximate information state model which learns a representation of the unknown impact of \CAV{1}'s actions on the actions of \HDV{2}. This is illustrated in Fig. \ref{fig:encode_decoder_with_action}, where we denote the encoder by $\psi: \mathcal{S} \times \mathcal{Y} \times \mathcal{U} \to \mathcal{S}$ and the decoder by $\phi: \mathcal{S} \times \mathcal{U} \to \mathcal{Y}^3$, where $\mathcal{Y} \subset \mathbb{R}^6$.
We seek to learn this approximate information state model purely from observation data. For this, we compare the distributions predicted by the decoder against data that constitutes sampled points from the ground truth. Next, we describe each component of our neural network architecture.

%%%%%%%%%%%%%%%%%%%%%%%%%%%%%%%%%%%%%%%%%%%%%%

\subsection{Encoder}

The encoder $\psi$ aims to learn a functional mapping which enforces $(\textbf{AP2a})$, i.e., its aim is to compress the history of observations into an approximate information state which can represent the \emph{CAV-HDV interactions}. Thus, at each $t\in \mathbb{N}$, the output of the encoder is the approximate information state $s_t = \sigma_t(m_t)$, where $m_t$ is the memory, and its input includes the previous approximate information state, latest observation, and previous control action given as a tuple $(s_{t-1}, y_t, u_{t-1})$. We illustrate this in Fig. \ref{fig:encode_decoder_with_action}. The encoder comprises of two fully connected layers followed by a recurrent neural network (RNN). The hidden state of the RNN at each $t\in \mathbb{N}$ is treated as the output of the encoder $s_t$. The linear layers are responsible for receiving the inputs $(y_t, u_{t-1})$, and the RNN ensures that the encoder at time $t$ receives the previous approximate information state $s_{t-1}$ as an input too. Thus, the structure of the RNN ensures that the approximate information state is updated at each $t\in \mathbb{N}$ as
   \begin{equation} \label{eq:encoder_update}
    s_{t} = \psi(s_{t-1}, y_{t}, u_{t-1}),
   \end{equation}   
where $\psi$ is a deterministic function and denotes the mapping generated by layers of neural networks within the encoder.

\subsection{Decoder}

%%%%%%%%%%% NOTE %%%%%%%%%%%%%%%%%%
% Specify why we are not using a DP and what alternative are we proposing (MPC) and refer it back to the problem formulation

 The decoder $\phi$ aims to learn a functional mapping which enforces $(\textbf{AP1})$ and $(\textbf{AP2b})$, i.e., its aim is to learn to predict distribution on future observations and the expected cost as a function of the approximate information state and the control action at any instance of time. In our implementation, the decoder $\phi$ comprises of three fully connected layers. At each $t\in \mathbb{N}$, the decoder takes as an input the approximate information state $s_t$ and control action $u_{1,t}$ of \CAV{1}. The decoder generates as an output a distribution of future states of \HDV{2} over a horizon $\III_t$, $(x_{2, t+1}, x_{2, t+2}, \dots, x_{2, t+H})$, rather than just the next observation $y_{t+1}$. Thus, the output of the decoder is given by
   \begin{equation} \label{eq:decoder_update}
    \phi({s}_{t},{u}_{1,t}) = \hat{p}^{\text{y}}({x}_{2,t+1},\dots, {x}_{2,t+H}~|~{s}_{t},{u}_{1,t}),
   \end{equation} 
where $\hat{p}^{\text{y}}$ is the predicted probability distribution on the observations. Thus, in our architecture, the decoder aims to enforce $(\textbf{AP2b})$ not just for time $t+1$ as described in \eqref{eq:ap_2b} but for the entire control horizon $\III_t$. This serves two purposes: (i) it allows for a planning depth across the control horizon $\III_t$ during the implementation of the MPC controller, and (ii) it focuses the output of the decoder on predicting the future states of \HDV{2}. 
We do not include an expected cost term in the output of the decoder because the cost \eqref{eq:mpc-obj} at each time is a deterministic function of the observations and predictions of \CAV{1}. 
Note that we can focus only on \HDV{2} because the dynamics of \CAV{1} are already known from \eqref{eq:integrator} and need not be learned. 
Thus, if the decoder learns to best predict the trajectories, it automatically ensures both $(\textbf{AP1})$ and $(\textbf{AP2b})$.

%%%%%%%%%%%%%%%%%%%%%%%%%%%%%%%%%%%%%%%%%%%%%%
\subsection{Training}
In this subsection, we describe how we train both the encoder $\psi$ and the decoder $\phi$ by considering the left-hand side in \eqref{eq:ap_2b} from ($\textbf{AP2b}$) as a training loss. Thus, the training loss measures the distance between a predicted probability distribution on observed future states and the underlying distribution on the ground truth using an integral probability metric.
For this purpose, we use the distance-based maximum mean discrepancy (MMD) metric defined in \cite[Proposition 32]{subramanian2022approximate} to characterize the distance between two probability distributions. 

Let the underlying distribution on the ground truth be denoted by $p$ and the predicted distribution of the decoder be $\hat{p}^{\text{y}}$. 
The training loss is then given by $d(\hat{p}^{\text{y}}, p)^2$, where $d(\cdot, \cdot)$ is the MMD metric.
From \eqref{eq:decoder_update}, note that the distribution $\hat{p}^{\text{y}}$ is parameterized by the weights of the encoder-decoder network. We consider that this is a multivariate normal distribution on the possible future realizations of the states $({x}_{2,t+1},\dots, {x}_{2,t+H})$ of \HDV{2} over the horizon $\III_t$. Then, we fix the standard deviation of $\hat{p}^{\text{y}}$ to $1$ and consider that outputs of the decoder represent the mean value vector of the distribution. At each $t\in \mathbb{N}$, we have access to a sampled observation $({x}^{\text{ref}}_{2,t+1}, \dots, {x}^{\text{ref}}_{2,t+H})$ from the underlying distribution $p$. To minimize the loss $d(\hat{p}^{\text{y}}, p)^2$, we need to construct an unbiased estimator of the gradient of this training loss using only sampled points. From  \cite[Proposition 33]{subramanian2022approximate}, the gradient $\nabla d(\hat{p}^{\text{y}}, p)^2$ can be estimated in an unbiased manner from the mean of $\hat{p}^{\text{y}}$ and sampled points from $p$ by the gradient of
%For each $t+i$, $i \in \III_t$, we compare the mean ${x}_{2,t+i}$ to a sample ${x}^{ref}_{2,t+i}$ from the true distribution $p$. To train the $\psi-\phi$ neural network structure, we need the gradient of the integral probability metric between the parameterized distribution and the true distribution. The gradient $\nabla d(\hat{p}^{\text{y}}, p)$ at each time $t+i$ over the horizon $\III_t$ can be be estimated in an unbiased manner by:
   \begin{equation} \label{eq:MMD}
   ( {x}_{2,t+i} -2\cdot{x}^{ref}_{2,t+i})^{\top} x_{2,t+i}. 
   \end{equation}   
%Note that \eqref{eq:MMD} is the gradient of the mean square error (MSE) loss between the predicted mean of states for \HDV{2} and the sample from the true distribution which is the ground truth over the horizon $\III_t$. Thus, we use the mean squared error
We use \eqref{eq:MMD} as a surrogate loss for the training, which is computed using the predicted mean $({x}_{2,t+1},\dots, {x}_{2,t+H})$ and the sample from the ground truth $({x}^{\text{ref}}_{2,t+1}, \dots, {x}^{\text{ref}}_{2,t+H})$.

\section{Simulation and Results}\label{section:Simulation}

%%%%%%%%%%%%%%%%%%%%%%%%%%%%%%%%%%%%%%%%%%%%%%
\subsection{Datasets}\label{subsection:Datasets}

We train and validate the use of approximate information state models for three datasets that contain trajectories of HDVs in highway on-ramp merging scenarios. 
For the first dataset, we collect trajectory data from the NGSIM repository for I-80 \cite{NGSIM}. Specifically, we extract the positions, velocities, and accelerations of the vehicle on the ramp as well as the two consecutive highway vehicles that it interacts with during merging.  
We use this data to validate the predictive ability of our approximate information state model for the trajectories of human-driven vehicles in real-life merging scenarios. The training details for specializing in the architecture from Section \ref{section:model} to perform this task and the results are presented in Subsection \ref{subsection:ngsim_results}.

Next, we use an \emph{inverse reinforcement learning} (IRL) technique described in \cite{kuderer_learning_2015} to replicate human drivers in simulations and generate two datasets for the scenario presented in Section \ref{section:Problem Formulation}.
With the IRL technique, we assume that the control actions of an HDV can be represented by the minimizer of a feature-based objective function in which different driving styles are characterized by the weights of the features.
By conducting and collecting data from simulations with different parameters for HDV's model, we can capture a full spectrum of possible actions for a human driver, ranging from conservative to aggressive driving.
This approach yields the second and third datasets used in the training and validation of approximate information state models in our subsequent exposition.
In the second dataset, we consider that \CAV{1} follows a safe control strategy proposed in \cite{Le2023ACC} against different HDVs. This ensures that only $1$\% of the situations included in the dataset demonstrate an unsafe merging scenario. % while the CAV's control actions selects actions using the control strategy proposed in \cite{Le2023ACC}.
%attempt to maximize safety and energy efficiency. 
In the third dataset, we consider that the control actions of both \HDV{2} and \CAV{1} are generated using the IRL technique with randomized weights. This simulates an exploratory control strategy for the CAV and does not include any consideration for safety during data generation.
We use the second and third datasets to validate the performance of the MPC strategy presented in Subsection \ref{subsection:IMPC}. This MPC approach utilizes the predictions of an approximate information state model trained individually on each dataset to generate control actions for \CAV{1} against a variety of behaviors for \HDV{2}.

%using two different policies for the merging scenario which was introduced in section \ref{section:Problem Formulation}.
%The first strategy used to generate dataset for the interaction between \CAV{1} and \HDV{2} is a cooperative, safe, and energy efficient strategy where the control actions of \CAV{1} are optimized to improve safety and energy efficiency. The second dataset is generated by a random strategy where the control actions for \CAV{1} need not be safe, energy optimal and cooperative with \HDV{2}. 

In the following subsections, we present the details for training an approximate information state model on each dataset and the validation of the trained models. Note that the encoder-decoder neural network architectures do not assume prior knowledge of the strategy or the model used to generate the data.

%%%%%%%%%%%%%%%%%%%%%%%%%%%%%%%%%%%%%%%%%%%%%%
\subsection{Network Architecture and Results for NGSIM} 
\label{subsection:ngsim_results}
\begin{figure*} 
\centering
\begin{subfigure}{.32\textwidth} \hspace{-10pt}
\includegraphics[width=1\textwidth]{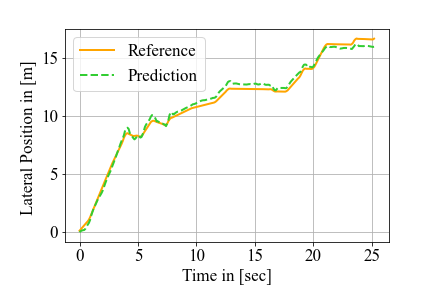}
%\caption{}
\end{subfigure} 
\begin{subfigure}{.32\textwidth} \hspace{-10pt}
\includegraphics[width=1\textwidth]{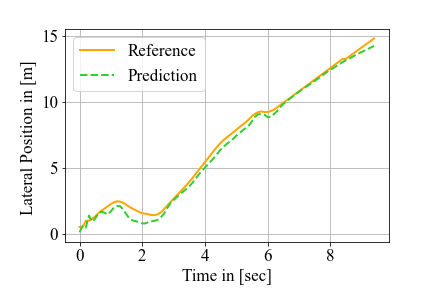}
%\caption{}
\end{subfigure}
\begin{subfigure}{.32\textwidth} \hspace{-10pt}
\centering
\includegraphics[width=1\textwidth]{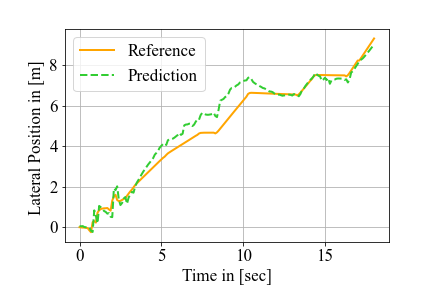}
%\caption{}
\end{subfigure} 
%\caption{}
\begin{subfigure}{.32\textwidth} \hspace{-10pt}
\includegraphics[width=1\textwidth]{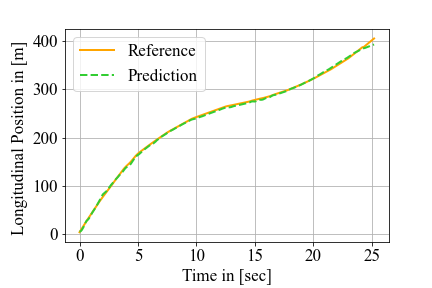}
\caption{}
\end{subfigure} 
\begin{subfigure}{.32\textwidth} \hspace{-10pt}
\includegraphics[width=1\textwidth]{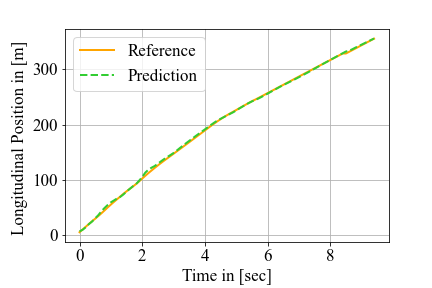}
\caption{}
\end{subfigure}
\begin{subfigure}{.32\textwidth} \hspace{-10pt}
\centering
\includegraphics[width=1\textwidth]{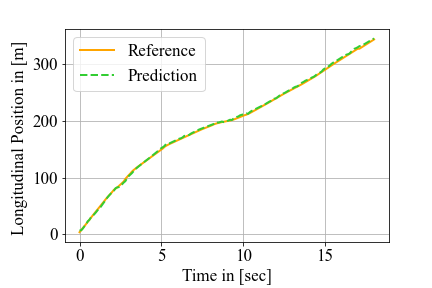}
\caption{}
\end{subfigure} 
\caption{A collection of plots showing the results of trajectory prediction v/s the reference trajectory from the NGSIM dataset for three different cars (a), (b), and (c) merging from the ramp.}
\label{fig:trajectory_results_NGSIM}
\vspace{-10pt}
\end{figure*}

In this subsection, we train both an encoder and a decoder to cumulatively learn an approximate information state model for the first dataset collected from the NGSIM repository. Recall from Subsection \ref{subsection:Datasets} that our dataset includes the trajectory information for the vehicle on the ramp and two consecutive highway vehicles that interact with it. 
Thus, at each instance of time, the encoder $\psi$ takes as input the most recent observation of each vehicle's state and action, i.e., a $9$ dimensional input consisting of a $2$ dimensional observation of position and speed and $1$ dimensional action for each vehicle. The encoder comprises two fully connected layers of dimensions $(9,8)$ and $(8,16)$ each with a ReLU activation, followed by a recurrent gated unit (GRU) \cite{cho2014properties} layer with a hidden state of size $24$. We consider the hidden state to be the output of the encoder and treating it as the approximate information state at any instance of time. The approximate information state and the most recent actions of each vehicle are provided as input to the decoder $\phi$. The decoder comprises three fully connected layers with dimensions $(24+3, 32), (32, 64)$, and $(64, 2)$, respectively, and with ReLU activation for the first two layers. The decoder is trained to predict the trajectory of the on-ramp vehicle for a horizon of $H=1$ time steps. A comparison of the predicted trajectories for three on-ramp vehicles with their actual trajectories is plotted in Fig. \ref{fig:trajectory_results_NGSIM}. Each column in this plot corresponds to a single vehicle, where the plots in the first row are a comparison of the position in the direction perpendicular to the highway (labeled as lateral position), and in the second row, we plot the position along the direction of the highway (labeled as longitudinal position). The approximate information state model performs well in predicting the longitudinal positions. At the same time, our model is capable of closely predicting the trajectory of the on-ramp vehicle along both the lateral and longitudinal axes. % The maximum error across both axes of each plotted trajectory is $0.5$ m for Fig. \ref{fig:trajectory_results_NGSIM}(a), $0.7$ m for Fig. \ref{fig:trajectory_results_NGSIM}(b), and $0.9$m for Fig. \ref{fig:trajectory_results_NGSIM}(c). 
The root mean square error between the reference and predicted trajectories in the lateral positions are $0.35$ m for \ref{fig:trajectory_results_NGSIM}(a), $0.39$ m for \ref{fig:trajectory_results_NGSIM}(b), and $0.54$ m for \ref{fig:trajectory_results_NGSIM}(c), whereas it is $2.62$ m for \ref{fig:trajectory_results_NGSIM}(a), $1.92$ m for \ref{fig:trajectory_results_NGSIM}(b), and $1.73$ m for \ref{fig:trajectory_results_NGSIM}(c) in the longitudinal positions. These results validate the predictive capability of our approximate information state model for real-life merging scenarios involving human-driven vehicles.

%%%%%%%%%%%%%%%%%%%%%%%%%%%%%%%%%%%%%%%%%%%%%%
\subsection{Network Architecture for IRL }

In this subsection, we present the details of the encoder-decoder architecture used to learn approximate information state models for both datasets generated using IRL.  

%Recall that the simulation based datasets were generated to train and validate the approximate information state models by using them in our MPC strategy presented in the next subsection. From here on we set the control horizon length $H=10$ for the purposes of training and control. We train an encoder-decoder architecture to learn an approximate information state representation for predicting the trajectory of \HDV{2}.

\begin{figure}[ht!]
    \vspace{-10pt}
  \centering
  %\captionsetup{justification=centering}
  \includegraphics[width= 1\columnwidth]{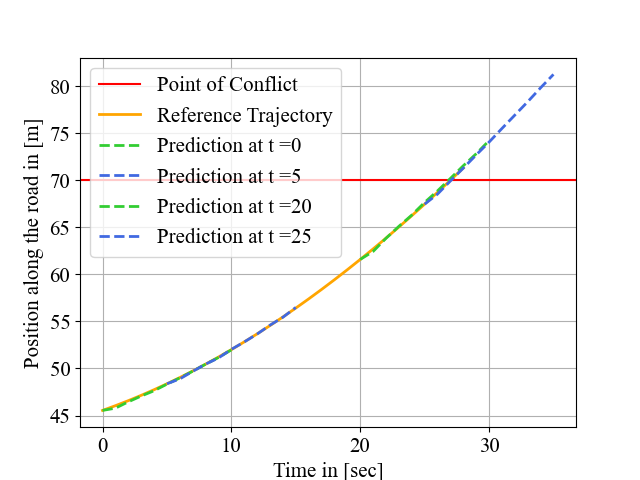}
  \caption{The results of trajectory prediction $v/s$ are the actual trajectory from the generated dataset.}
  \label{fig:trajectory_results_IRL}
  \vspace{-5pt}
\end{figure}

The encoder and decoder follow the same underlying structure as the networks used in Subsection \ref{subsection:ngsim_results}. Since we consider the scenario presented in Section \ref{section:Problem Formulation}, we reduce the number of inputs and outputs in each layer compared to the previous networks. Note that we use identical networks across the generated datasets. At each instance of time, the encoder $\psi$ takes as input the most recent observation of the state and action of both \CAV{1} and \HDV{2}, which is a $6$ dimensional input consisting of a $2$ dimensional observation of position and speed and $1$ dimensional action for each vehicle. The encoder structure is given by $(6, 8), (8, 16)$ with ReLU activation followed by a GRU with a hidden state of size $4$. Once again, we consider the hidden state to be the output of the encoder and treating it as the approximate information state at any instance of time. The approximate information state and the most recent action of \CAV{1} is provided as input to the decoder $\phi$. The decoder comprises three fully connected layers of dimensions $(4+1, 2), (2, 4)$, and $(4, H)$, respectively,  where H is the length of the horizon and a ReLU activation for the first two layers.
%The output of the decoder is a prediction of the future trajectory of \HDV{2}
We use the distance-based MMD described in \eqref{eq:MMD} to train the encoder-decoder neural network architectures. We present the results of trajectory prediction in Fig. \ref{fig:trajectory_results_IRL}, where we compare the predicted trajectories at different time instances $t=0,5,20,25$ sec to the reference trajectory. The root mean square error between all predicted and reference trajectories is $0.17$ m. For these datasets, we set the length of the control zone $L_c=70$ m and we indicate the conflict point using a red horizontal line.   

%where trajectories for \HDV{2} are predicted only after \CAV{1} has entered the control zone \emph{i.e.} \HDV{2} could enter need not be at the beginning of the control zone when 

\subsection{Iterative MPC Implementation}\label{subsection:IMPC}

To solve the MPC problem and obtain the control input $u_{1,t}$ for \CAV{1} at the current time step $t$, we utilize the encoder-decoder neural network to predict the human driving trajectory over the control horizon.
Meanwhile, the decoder of the neural network also requires $u_{1,t}$ as the input to make the prediction.
One can consider the encoder-decoder neural network directly as a constraint in the MPC problem.
However, due to the complexity of the neural network model, the resulting MPC problem becomes computationally intractable.
Therefore, we propose an iterative MPC implementation that sequentially computes the neural network prediction and solves the MPC problem.
The iterative MPC implementation can be detailed in Algorithm~\ref{alg:iterative}.
At each time step, we initialize $u_{1,t}^{(0)} = 0$ and compute the prediction of the approximate information state by ${s}_t = \psi({s}_{t-1}, {x}_t, {u}_{t-1})$ where the inputs include the previous approximate information state ${s}_{t-1}$, current states ${x}_t$, and previous control inputs ${u}_{t-1}$.
Next, at each iteration $j$ of the algorithm, we predict states and control actions of HDV–2 by ${y}_{t+1:t+H}^{(j)}=\phi({s}_{t}, {u}_{1,t}^{(j-1)})$ that use the approximate information state ${s}_t$ and the control input ${u}_{1,t}^{(j-1)}$ of the MPC in the last iteration.
Then we solve the MPC problem given the prediction of ${y}_{t+1:t+H}^{(j)}$ to obtain the new control inputs ${u}_{1,t}^{(j)}$.
This procedure is repeated until a maximum number of iterations $j_{\mathrm{max}}$ is reached.

\begin{algorithm}[ht!]
  \caption{Iterative MPC Implementation}
  \label{alg:iterative}
  \begin{algorithmic}[1]
    \Require $t$, $H \in \NN$, $j_{\mathrm{max}} \in \NN$, $u_{1,t}^{(0)} = 0$ 
    \State Predict the approximate information state by ${s}_t = \psi({s}_{t-1},{x}_t,{u}_{t-1})$
    \For{$j = 1,2,\dots,j_{\mathrm{max}}$}
    \State Predict states and control actions of \HDV{2} by ${y}_{t+1:t+H}^{(j)}=\phi({s}_{t}, {u}_{1,t}^{(j-1)})$  
    \State Solve \eqref{eq:MPC} to obtain ${u}_{1,t:t+H-1}^{(j)}$
    \EndFor
    \State \Return $u_{1,t}^{(j_{\mathrm{max}})}$
  \end{algorithmic}
\end{algorithm} \setlength{\textfloatsep}{0.15cm}

\vspace{-10pt}
%%%%%%%%%%%%%%%%%%%%%%%%%%%%%%%%%%%%%%%%%%%%%%
% Viet-Anh is writing this subsection
\subsection{Simulation Results}

We conducted simulations in Python programming language in which CasADi \cite{andersson_casadi_2019} and the built-in IPOPT solver \cite{wachter_implementation_2006} 
are used for formulating and solving the MPC problem, respectively.
The parameters of MPC were chosen as:
${\Delta T = \SI{0.2} {s}}$,
${H = 10}$, 
${v_\mathrm{min} = \SI{0.0}{m/s}}$, 
${v_\mathrm{max} = \SI{14.0} {m/s}}$,
${u_\mathrm{min} = \SI{-3.0} {m/s^2}}$,
${u_\mathrm{max} = \SI{2.0} {m/s^2}}$, 
${\omega_1 = 1.0}$, 
${\omega_2 = 10.0}$, 
${\omega_3 = 10^3}$, 
${\rho = \SI{1.0} {s}}$.
Recall that we simulate human driving behavior by using an IRL-based model with different weights to validate the control method with different driving styles.

\begin{figure*}
\centering
\begin{subfigure}{.32\textwidth} \hspace{-10pt}
\includegraphics[width=1\textwidth]{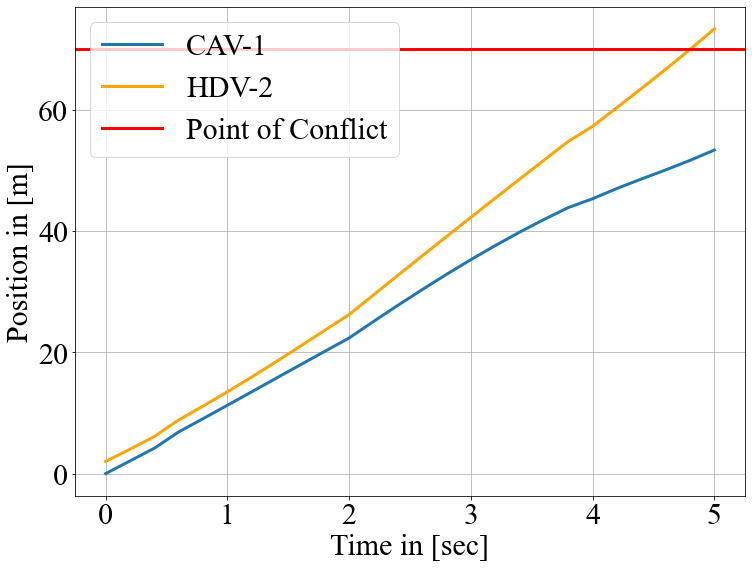}
\caption{}
\end{subfigure} 
\begin{subfigure}{.32\textwidth} \hspace{-10pt}
\includegraphics[width=1\textwidth]{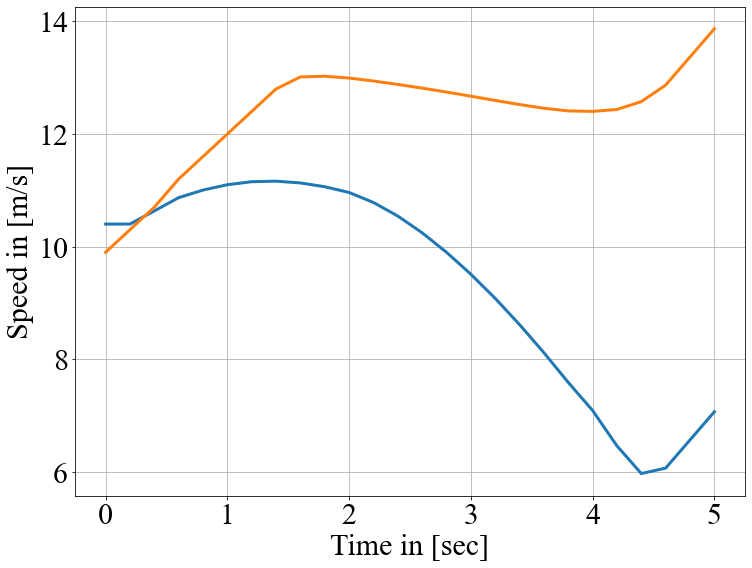}
\caption{}
\end{subfigure}
\begin{subfigure}{.32\textwidth} \hspace{-10pt}
\centering
\includegraphics[width=1\textwidth]{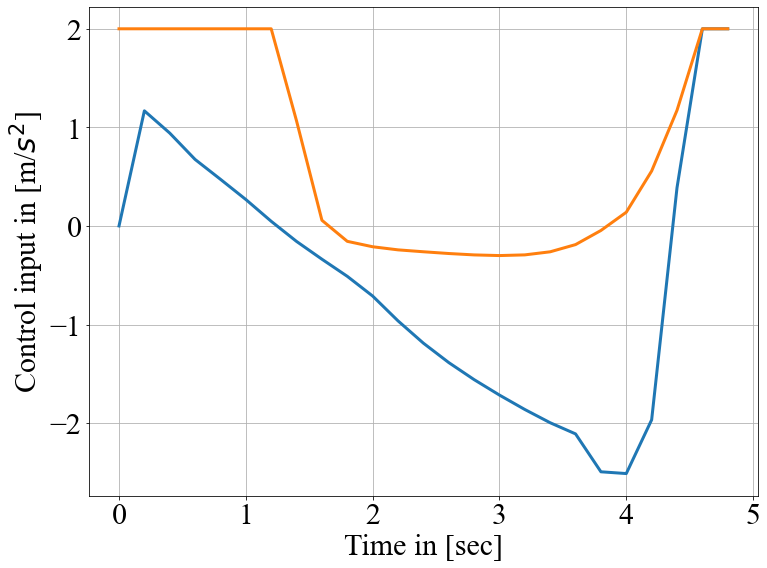}
\caption{}
\end{subfigure} 
\caption{Trajectories, speeds, and distances of the vehicles in the simulation with an aggressive human driver.}
\label{fig:traj1}
\end{figure*}

\begin{figure*}
\centering
\begin{subfigure}{.32\textwidth} \hspace{-10pt}
\includegraphics[width=1\textwidth]{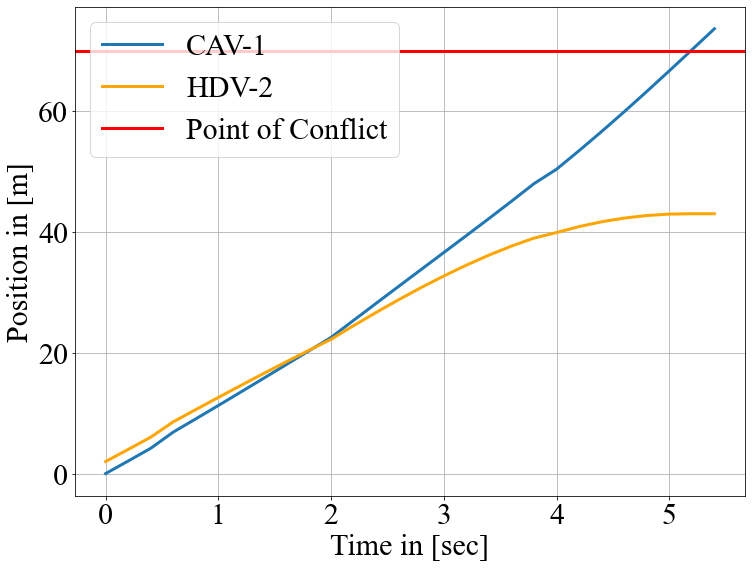}
\caption{}
\end{subfigure} 
\begin{subfigure}{.32\textwidth} \hspace{-10pt}
\includegraphics[width=1\textwidth]{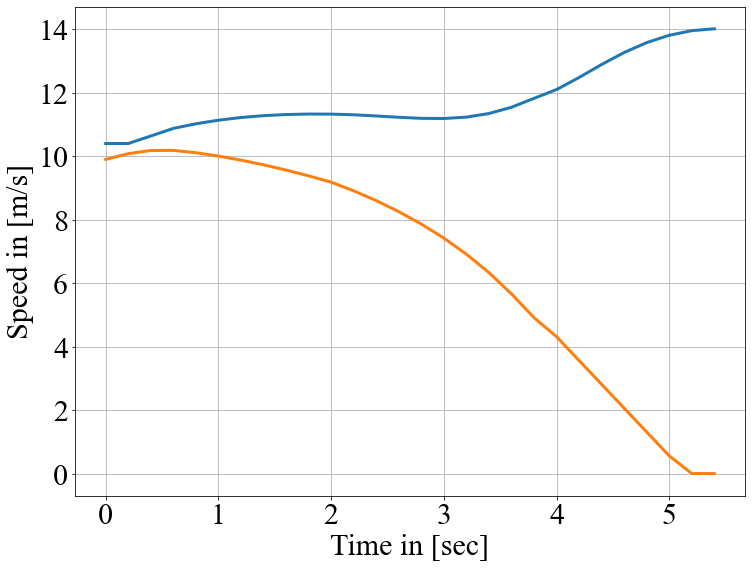}
\caption{}
\end{subfigure}
\begin{subfigure}{.32\textwidth} \hspace{-10pt}
\centering
\includegraphics[width=1\textwidth]{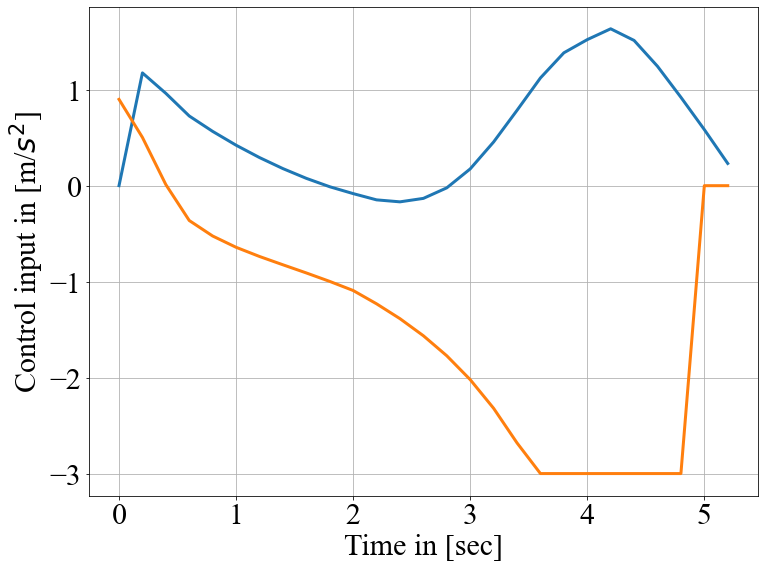}
\caption{}
\end{subfigure} 
\caption{Trajectories, speeds, and distances of the vehicles in the simulation with a conservative human driver.}
\label{fig:traj2}
\end{figure*}

Figures~\ref{fig:traj1} and \ref{fig:traj2} show the trajectories, speeds, and control inputs of \CAV{1} and \HDV{2} in two simulations with two human driving styles, \ie an aggressive and a conservative human driver.
As shown from the figures, in both simulations, safe distance is guaranteed, but \CAV{1} has different merging maneuvers depending on the behavior of \HDV{2}.
If the human driver accelerates to cross first, then \CAV{1} slows down to yield, while if the human driver is more conservative, \CAV{1} finds it safe to cross before \HDV{2}.

\begin{table}[!ht]
\vspace{-5pt}
  \caption{Comparison between models trained using safe and exploratory strategy data.}
  \label{tab:compare} 
  \centering
  \begin{tabular}{ c | c c }
    \toprule[1pt]% <-- Toprule here
    \textbf{Reaction Time Delay $(\rho)$ } & \textbf{Safe } & \textbf{Exploratory} \\
    \midrule[0.5pt] % <-- Midrule here
     0.6 & 4984 (99.6\%) & 4578 (91.5\%)\\ 
     0.8 & 4996 (99.9\%) & 4793 (95.8\%)\\
     1.0 & 5000 (100\%) & 4971 (99.4\%)\\
    \bottomrule[1pt] % <-- Bottomrule here
  \end{tabular}
  \vspace{-5pt}
\end{table}

To evaluate the performance of the MPC with two learned prediction models, we conduct 5000 simulations with different initial conditions of the vehicles and different IRL models for \HDV{2} and compare the level of safety under three different values for $\rho$. Recall from Subsection \ref{subsection:mpc_formulation} that \CAV{1}'s conservatism increases with $\rho$. Hence, we observe that the number of unsafe situations decreases with an increase in $\rho$.
The results are indicated in Table~\ref{tab:compare} and show that MPC with the model trained using the safe-strategy dataset achieves a higher level of safety compared to the model trained using the exploratory-strategy dataset. We believe that this might be due to the model for the exploratory dataset requiring further training.
 
%\begin{table}[!bt]
%  \caption{Comparison between weight adaptation strategies using Bayesian optimization (BayOpt) and SVO.}
%  \label{tab:compare} 
%  \centering
%  \begin{tabular}{ p{0.22\textwidth} | p{0.095\textwidth} p{0.095\textwidth} }
%    \toprule[1pt]% <-- Toprule here
%    \textbf{Comparison metrics} & \textbf{BayOpt} & \textbf{SVO} \\
%    \midrule[0.5pt] % <-- Midrule here
%    Number of simulations with safety &  & \\
%    Number of simulations with time improvement & & \\
%    \bottomrule[1pt] % <-- Bottomrule here
%  \end{tabular}
%  \vspace{-5pt}
%\end{table}

\section{Concluding Remarks}\label{section:conclusion}

In this paper, we presented an approach to learn an approximate information state model of CAV-HDV interactions for a CAV to maneuver safely during highway merging. We validated the capability of our approximate information state model by training with and predicting real-life merging scenarios involving human-driven vehicles for the NGSIM repository. Then, we showed that using data generated from an IRL model for a mixed traffic scenario with a CAV and an HDV, an approximate information state model can be learned to predict future trajectories of the HDV. We proposed an iterative MPC algorithm to generate control actions for the CAV utilizing these predictions and showed using numerical simulations that they are safe against a spectrum of driving behaviors of the HDV. 
%numerically simulate highway merging scenarios and generated safe control policies using iterative MPC for a CAV to merge with HDVs that demonstrate a spectrum of driving behaviors and a set of iterative MPC parameters related to safety. 
Future research should focus on (i) extending the framework for continuous-time control, (ii) considering a more diverse range of scenarios with multiple lanes, multiple cars, roundabouts, and intersections, and (iii) investigating the practical effectiveness of the
a proposed approach using an experimental testbed \cite{chalaki2022research}.

\bibliographystyle{ieeetr}
\bibliography{References,Latest_IDS,references_VA}

\end{document}